\documentclass[runningheads]{llncs}

\usepackage{graphicx}
\usepackage{comment}
\usepackage{amsmath,amssymb}
\usepackage{color}
\usepackage{url}
\usepackage{hyperref}
\usepackage{cleveref}
\usepackage{booktabs}
\usepackage{subcaption}
\usepackage{xspace}
\usepackage{pifont}
\usepackage{soul}

\newcommand{\cmark}{\ding{51}}%
\newcommand{\xmark}{\ding{55}}%
\newcommand{\red}[1]{{\color{black}#1}}
\newcommand{\blue}[1]{{\color{black}#1}}

\newcommand{\MethodName}{\mbox{UVOTE}\xspace}

\newif\ifreview
\reviewfalse

\ifreview
	\usepackage{lineno}

	\linenumbers
\fi

\begin{document}


\def\SubNumber{75}

\def\GCPRTrack{Fast Review Track}

\title{Uncertainty Voting Ensemble for Imbalanced Deep Regression}

\ifreview
\else

	
	
        \author{Yuchang Jiang\inst{1} \and
        Vivien Sainte Fare Garnot\inst{1} \and
        Konrad Schindler\inst{2} \and
        Jan Dirk Wegner\inst{1} }
        
        \authorrunning{Y.~Jiang et al.}
        
        \institute{University of Zurich, Zurich, Switzerland \and
        ETH Zurich, Zurich, Switzerland \\
        \email{yuchang.jiang,vsaint,jandirk@uzh.ch}\\
        \email{schindler@ethz.ch}}
\fi

\maketitle              

\begin{abstract}

  Data imbalance is ubiquitous when applying machine learning to real-world problems, particularly regression problems. If training data are imbalanced, the learning is dominated by the densely covered regions of the target distribution and the learned regressor tends to exhibit poor performance in sparsely covered regions.
Beyond standard measures like oversampling or reweighting, there are two main approaches to handling learning from imbalanced data. For regression, recent work leverages the continuity of the distribution, while for classification, the trend has been to use ensemble methods, allowing some members to specialize in predictions for sparser regions.
In our method, named \MethodName, we integrate recent advances in probabilistic deep learning with an ensemble approach for imbalanced regression. We replace traditional regression losses with negative log-likelihood, which also predicts sample-wise aleatoric uncertainty. Our experiments show that this loss function handles imbalance better. Additionally, we use the predicted aleatoric uncertainty values to fuse the predictions of different expert models in the ensemble, eliminating the need for a separate aggregation module.
We compare our method with existing alternatives on multiple public benchmarks and show that \MethodName consistently outperforms the prior art, while at the same time producing better calibrated uncertainty estimates. Our code is available at \hyperlink{https://github.com/SherryJYC/UVOTE}{https://github.com/SherryJYC/UVOTE}.

\end{abstract}
\section{Introduction}
\label{sec:intro}

Data imbalance is the norm, rather than the exception in real-world machine learning applications, and in regression tasks, in particular. Outside the realm of carefully curated research datasets, the distribution of the target values is typically non-uniform. Some parts of the distribution are covered by training examples much more densely than others, and as a result, machine learning models tend to be biased towards those well-represented regions and perform poorly in under-represented ones \cite{he2009learning}. What is more, these sparse regions of the distribution are often important. In several applications, the prediction results matter specifically for rare, unusual conditions like extreme wind speeds in meteorology \cite{maskey2020Winddeepti}, or particularly high biomass in vegetation mapping \cite{lang2022vhmglobalreg}. Therefore, addressing the imbalance problem is an active area of machine learning research.

\begin{figure}[]
  \centering
  \includegraphics[width=.5\textwidth, trim=.5cm 12cm 0.5cm 0cm, clip]{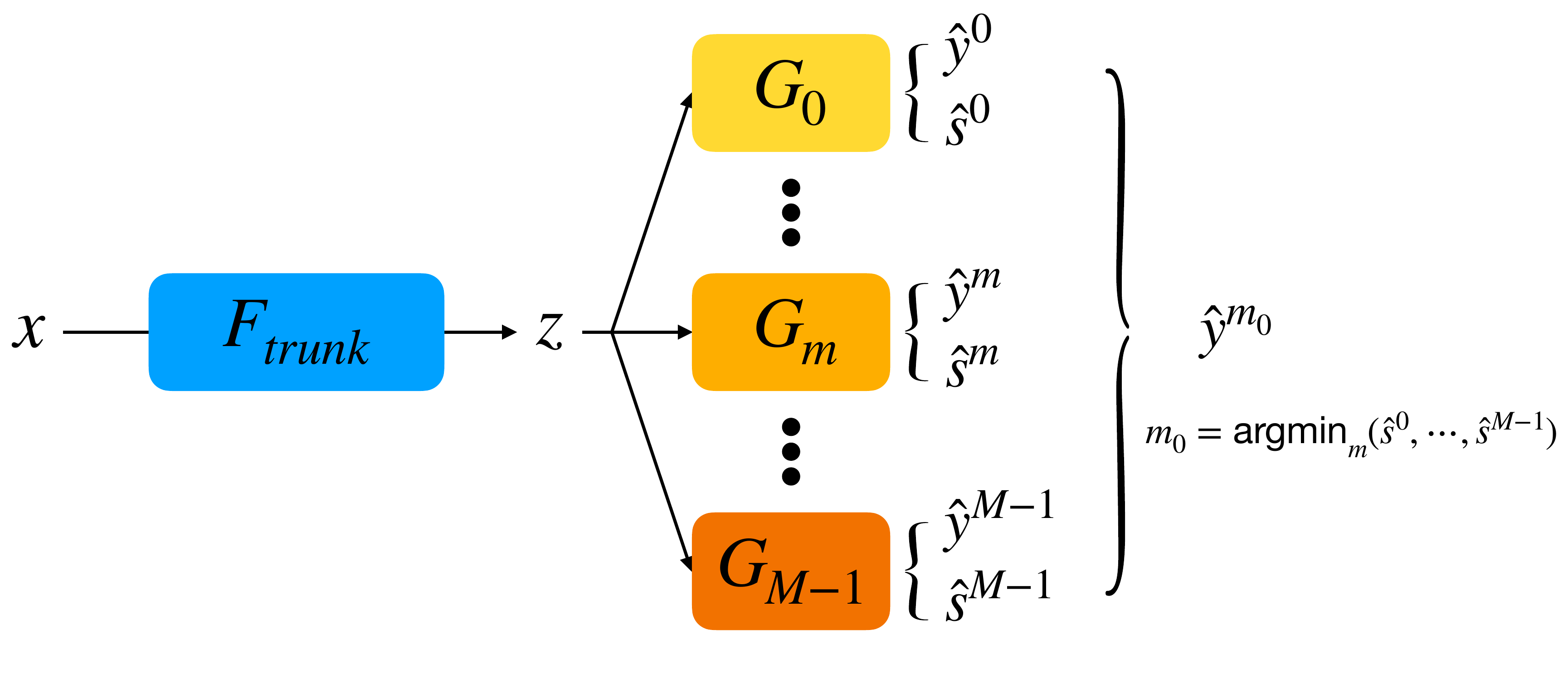}
  \vspace{-10pt}
  \caption{\textbf{Overview of \MethodName}. A shared backbone encodes the input $x$ into a representation $z$. A mixture of $M$ different experts uses this shared representation to make their predictions. Each expert predicts a regression value $\hat{y}$ as well as the uncertainty $\hat{s}$ of that prediction. At inference time, we use the prediction of the most certain expert $m_0$.}
  \label{model_overview}
\end{figure}

Traditional attempts to mitigate the impact of imbalance rely either on over-sampling rare data samples or on re-weighting the loss function to increase the cost of prediction errors at rare samples \cite{he2009learning}. More recently, several authors have revisited the issue in the context of deep learning, typically \red{through an ensemble framework}. An ensemble of "expert" models is trained in such a way that they can each \blue{specialise on a } different part of the distribution. Then their predictions are aggregated to obtain the final inference. The challenge in such methods consists in ensuring complementarity between the different experts and designing an aggregation method that synthesizes the predictions of individual ensemble members according to their relevance. A naive solution is to use the ensemble average, but this risks giving too much weight to predictions that are irrelevant to the specific data point. More elaborate solutions tune the aggregation weights in an unsupervised fashion \cite{zhang2021self}. Once optimized, these weights are still fixed and subject to the same limitation. It has also been proposed to use dynamic weights obtained from a \emph{sample-level} voting module that is trained with an independent objective \cite{wang2020ride}.
All works mentioned so far focus on classification problems. \emph{Imbalanced regression}, on the other hand, has been studied a lot less and has only recently started to gain attention, especially since the publication of a suitable benchmark \cite{yang2021delving}. The prevalent idea so far has been to exploit the continuity of regression functions, either by smoothing the features and labels \cite{yang2021delving} or by regularizers that encourage similar latent features at similar (continuous) labels \cite{gong2022ranksim}.
On the contrary, the \red{ensemble} idea has barely been explored in the context of imbalanced regression, despite the fact that model ensembles are common for deep regression \cite{lang2022vhmglobalreg,yeo2021robustness,becker2023country}.

Here, we introduce \red{\underline{U}ncertainty \underline{VOT}ing \underline{E}nsemble} for deep imbalanced regression (\MethodName). We adopt an ensemble framework for regression and propose a principled and straightforward way to dynamically aggregate the predictions. Rather than adding an empirically designed or learned voting module, we leverage the fact that uncertainty estimation techniques for deep regression \cite{kendall2017uncertainties} inherently compute statistically meaningful weighting coefficients. Specifically, we use the estimated aleatoric uncertainties of individual experts to combine their predictions. To achieve this with a low computational overhead, we follow recent literature \cite{zhou2020bbn} and construct a light ensemble, consisting of a shared encoder backbone and separate decoding branches for different experts. We experimentally evaluate our approach against other methods for deep imbalanced regression on a diverse set of tasks, including age regression, meteorological prediction, and text similarity prediction. \MethodName sets a new state-of-the-art across all four datasets.
Importantly, while \MethodName improves overall performance, the gains are most significant for rare output values that are under-represented in the training data.
As an additional benefit, the uncertainties predicted by \MethodName are better calibrated and, therefore, more informative for downstream tasks that rely on the regression output.
Following this approach, we integrate uncertainty estimates from experts who specialize in different data distributions to mitigate the impact of imbalanced data on regression. Our contributions are as follows:
\begin{itemize}
    \item We introduce \MethodName, a novel, efficient end-to-end method for imbalanced regression.
    \item \MethodName outperforms all competing methods on four challenging datasets. It has lower regression errors while delivering uncertainty estimates that are well calibrated .
    \item To the best of our knowledge, \MethodName is the first deep imbalanced regression method that leverages probabilistic deep learning for ensemble aggregation. 
\end{itemize}

\section{Related Work}
\label{sec:related}

\paragraph{\bf Imbalanced Regression} As imbalanced regression receives less attention than imbalanced classification, early works usually use methods originally proposed for imbalanced classification. For example, Synthetic Minority Oversampling Technique (SMOTE) introduced in \cite{chawla2002smote} can create synthetic samples to reduce the imbalance in classification, which is also applied in regression problems \cite{branco2017smognReg}. Similarly, \cite{lang2022vhmglobalreg} and \cite{steininger2021denseWeight} follow the class-balanced loss idea to add frequency-based weights to the loss function, so the model pays more attention to minority samples. 
More recently, \cite{yang2021delving} introduces a public benchmark for imbalanced regression and leverage the continuity of the regression targets with label and feature smoothing techniques. This public benchmark has encouraged more work focusing on the imbalanced regression. RankSim \cite{gong2022ranksim} proposes a ranking-based regularization method to also utilize the target continuity and enhance representation learning in the imbalanced regression. \cite{ren2022balancedMSE} introduces a novel loss function by combining the training label distribution prior with the conventional mean-square-error loss, effectively addressing the issue of imbalance in regression.

\paragraph{\bf Imbalanced Classification} Most works studying imbalanced datasets focus on classification tasks. Early works include re-weighting \cite{cui2019classbalance}, re-sampling, \cite{chawla2002smote} and data augmentation \cite{zhang2017mixup}. 
\red{
Another
direction in imbalanced classification explores ensemble approaches, training multiple members and aggregating their predictions. The main challenges in such approaches are to ensure member diversity and to effectively aggregate their predictions.  \cite{zhou2018diverse} address this by assigning different subsets of the training dataset to each model expert to ensure diversity and aggregating the predictions using a simple average.
\cite{wang2020ride} proposes a two-stage method: in the first stage, they optimize three experts as three branches and use Kullback–Leibler divergence in the loss function to encourage expert diversity; in the second stage, they aggregate experts by training binary classifiers as dynamic expert assignment modules. \cite{zhang2021self} enforces different distribution for each expert explicitly to ensure the diversity of experts and utilizes a self-supervised training method at the test stage to combine experts for the final output. Although \cite{zhang2021self} requires no additional training for expert aggregation, the learnt weights are fixed at the \emph{dataset-level} instead of \emph{sample-level}. 
\cite{wang2023diversity} proposes a two-stage probabilistic ensemble method for classification. Their method involves creating a Gaussian mixture model with a group of trained models to generate a new, distilled model.
In comparison, our work 
proposes an end-to-end ensemble method that uses sample frequency as the source of diversity and introduces a new uncertainty-based expert aggregation mechanism. This mechanism requires no additional training and combines experts based on per-sample weights.
}

\paragraph{\bf  Uncertainty Estimation}

Probabilistic deep learning methods like \cite{kendall2017uncertainties} estimate both the mean target value and the uncertainty, which helps model interpretation.
Recent works further use uncertainty to achieve stronger predictions instead of solely producing uncertainty as a nice-to-have output. \cite{yeo2021robustness} utilizes an ensemble of probabilistic deep learning models to increase robustness to domain shifts. Each member of the ensemble is uniquely perturbed during training, and the aggregated ensemble prediction
via the corresponding uncertainty achieves a more robust prediction against image corruption.
Similarly, \cite{becker2023country} combines the predictions based on uncertainty to generate the country-wide map of forest structure variables, which is more robust to clouds.
Here we also use estimated uncertainty to aggregate the ensemble's prediction. Our approach differs from previous methods in two aspects. 
First, we use multiple branches trained with different losses instead of a simple ensemble of models with different initializations, which is more computationally efficient.
\blue{Second, we actively promote diversity within our ensemble members by assigning different sample weighting functions to each member, so that they focus on different parts of the distribution.} 

\section{Methods}

Fig. \ref{model_overview} shows a schematic overview of \MethodName, the following paragraphs describe its components. \MethodName consists of joint training of $M$ different regression experts. Each expert predicts a sample-dependent aleatoric uncertainty, and that uncertainty is used to combine the predictions.

We consider a generic univariate regression dataset $\mathcal{D}=\{(x_n, y_n), n \in [\![ 1, N ]\!] \}$ of size $N$, with $x_n$ the input tensors, and $y_n$ the corresponding scalar target values. We define $B$ equally spaced bins across the target range and approximate the frequency distribution of the data by counting the number of data points per bin, $\mathbf{f} = (f_1, \cdots, f_B)$. 

\paragraph{\bf Multi-headed architecture} Instead of training $M$ independent models, we follow recent literature \cite{zhou2020bbn} and design a multi-headed architecture with a shared backbone encoder and $M$ regression heads that act as different experts. This design has the advantage that it is computationally lightweight and lets all experts rely on a common representation space. The shared backbone encoder $F_{trunk}$ can be selected according to the task at hand and maps each input point $x_n$ to an embedding  $z_n$. The latter is processed by $M$ different regression heads $G_m$ that each output their individual expert prediction.

\paragraph{\bf Aleatoric uncertainty prediction} 

Each expert $m$ makes two predictions: the target value $\hat{y}_n^m$ and its associated aleatoric uncertainty $\hat{s}_n^m$. Following \cite{yeo2021robustness}, we train these predictions by minimizing the negative log-likelihood of the Laplace distribution:

\begin{align}
    \hat{y}_n^m, \hat{s}_n^m & = G_m(z_n) \:, \\
    \mathcal{L}^m_{NLL} &= \frac{1}{N} \sum^N_{n=1} w^m_n \big(exp(-\hat{s}^m_n)|y_n - \hat{y}^m_n | + \hat{s}^m_n\big) \:.
    \label{eq:nll}
\end{align}

For numeric stability, we optimize $\hat{s}_n$, the logarithm of the scale parameter in the Laplace distribution. 

\paragraph{\bf Joint training of diverse experts} 
Each expert $m$ is trained with a different weighting of the samples $w_n^m$, so as to achieve diversity and to make experts focus on different parts of the target distribution. The weights for expert $m$ are defined as:

\begin{figure}[th]
    \begin{minipage}{0.3\linewidth}
        \begin{equation}
        \label{equation:weight_fre_bin}
            w^m_n = \left(\frac{1}{f_{b(n)}}\right)^{p_m} \:, 
        \end{equation}
    \end{minipage}
    \hfill
    \begin{minipage}{0.6\linewidth}
        \begin{equation}
            \text{with} \:\: p_m = \frac{m}{M - 1}, m \in \{0, ..., M-1\} \: ,
        \end{equation}
    \end{minipage}
\end{figure}

where $b(n)$ denotes the bin in which sample $n$ falls.
Parameter $p_m$ controls how strongly an expert concentrates on samples from sparse regions of the input distribution, with larger $p$ corresponding to stronger rebalancing: when $p = 0$, the expert treats each sample equally; when $p = 1$, the expert employs inverse-frequency weighting and fully compensates density variations in the input.
Different settings of $p$ are complementary: unweighted standard regression learns the correct frequency prior and gives all data points the same influence on the latent representation; whereas inverse frequency weighting ensures that the model is not dominated by the dense regions of the distribution and fails to learn about the sparse ones. Intermediate versions between those extremes, like  inverse-squareroot weighting~\cite{lang2022vhmglobalreg}, attempt to find a compromise.
The ensemble of experts strikes a balance by offering solutions according to several different weighting schemes and picking the least uncertain one on a case-by-case basis.

\paragraph{\bf Dynamic learning}
For representation learning, it is arguably more correct to assign samples equal weight. It is not obvious why the feature extractor that transforms raw data into a latent representation should to a large degree depend on the properties of rare, potentially not overly representative samples. Inspired by \cite{zhou2020bbn}, we employ a dynamic learning strategy that initially focuses on the latent encoding and gradually phases in the remaining experts that have unequal weighting schemes:


\begin{figure}[th]
    \begin{minipage}{0.45\linewidth}
        \begin{equation}
            \mathcal{L} = \alpha \mathcal{L}^0_{NLL} + (1-\alpha)\sum^{M-1}_{m=1} \mathcal{L}^m_{NLL} \:,
        \label{eq:dyl}
        \end{equation}
    \end{minipage}
    \hfill
    \begin{minipage}{0.45\linewidth}
        \begin{equation}
            \alpha = 1 - \left(\frac{T}{T_{max}}\right)^2 \:,
        \end{equation}
    \end{minipage}
\end{figure}

where $T$ is the current epoch number, and $T_{max}$ is the maximum epoch number. $\mathcal{L}^0_{NLL}$ is the loss for expert $m = 0$, which treats all samples equally. $\alpha$ balances representation learning against mitigating the data imbalance. 

\paragraph{\bf Uncertainty-based expert aggregation}

During inference time, the predictions from multiple experts are combined based on the estimated uncertainty. One natural solution would be to weight the predictions using the inverse uncertainties. However, we obtain better experimental performance with selecting the output with the lowest predicted uncertainty:



\begin{figure}[th]
    \begin{minipage}{0.3\linewidth}
        \begin{equation}
            \hat{y}_n  = \hat{y}_n^{m_0} \:, \:\: \hat{s}_n  = \hat{s}_n^{m_0} 
        \end{equation}
    \end{minipage}
    \hfill
    \begin{minipage}{0.6\linewidth}
        \begin{equation}
            \text{with} \: m_0 = \text{argmin}_m(\hat{s}^1_n, \cdots  ,\hat{s}^M_n) \:.
        \end{equation}
    \end{minipage}
\end{figure}
\section{Experiments}

\subsection{Datasets}
We evaluate our method on the four public regression datasets:
%
\textbf{AgeDB} \cite{moschoglou2017AgeDB,yang2021delving} is an age estimation dataset and it consists of 12208 training images, 2140 validation images, and 2140 test images. 
\textbf{IMDB-WIKI} \cite{rothe2018IMDB,yang2021delving} is an age dataset consisting of 191509 training samples, 11022 validation samples, and 11022 test samples.
%
\textbf{Wind} \cite{maskey2020Winddeepti} is a wind speed estimation dataset consisting of satellite images. It contains 54735 images for training, 14534 images for validation, and 44089 images for testing. The wind speed ranges from 0 to 185 $kn$. 
%
\textbf{STS-B} \cite{cer2017semantic_STS_B,wang2018glue_STS_B,yang2021delving} is a semantic textual similarity benchmark, consisting of sentence pairs with a similarity score ranging from 0 to 5. It contains 5249 training samples, 1000 validation samples, and 1000 test samples. The four datasets have diverse imbalance factors, ranging from 353 to 5400.

\begin{figure}[th]
     \centering
     \begin{subfigure}[b]{.23\textwidth}
         \centering
         \includegraphics[width=\textwidth, height=.75\textwidth]{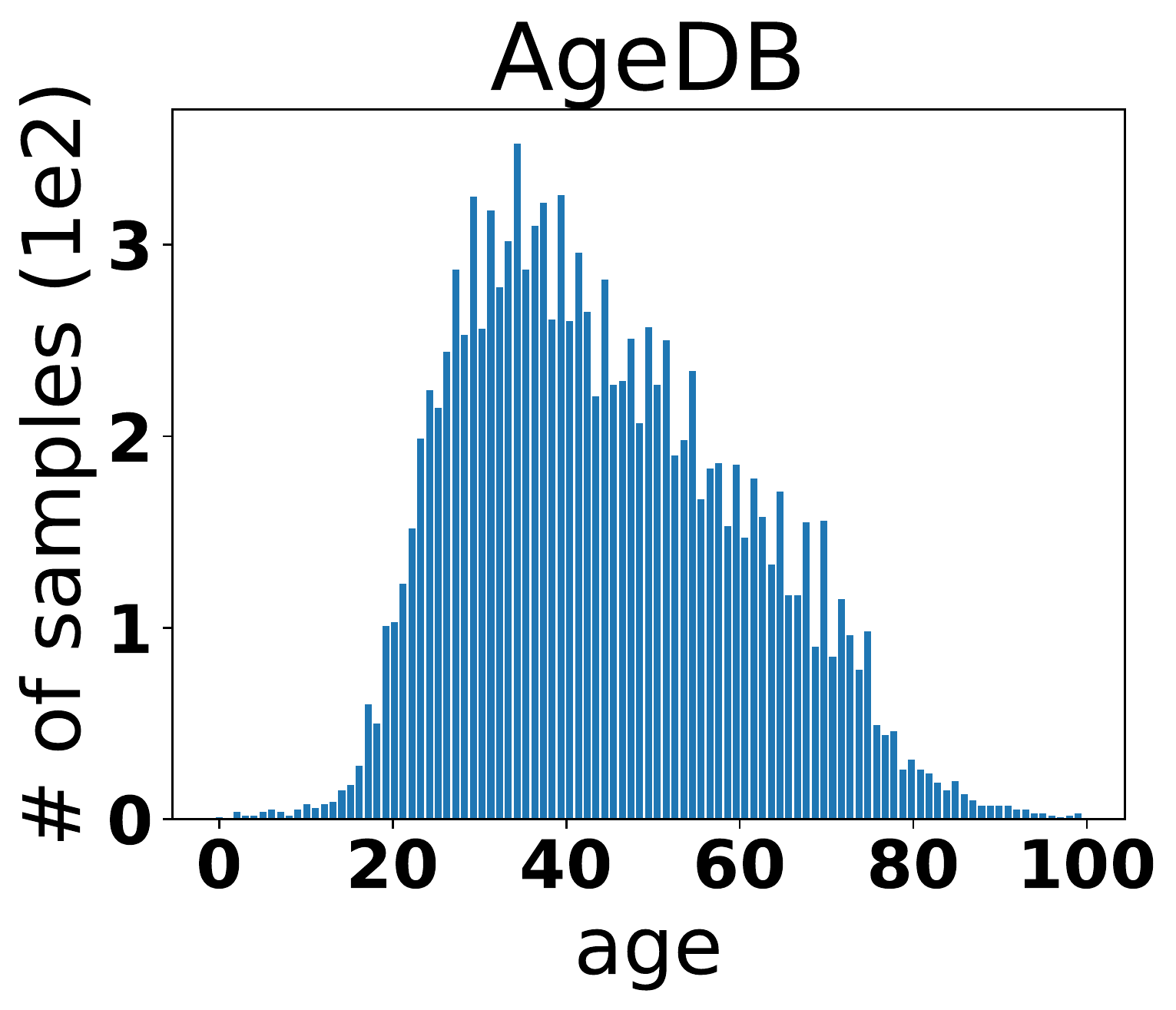}
         \label{fig:hist_agedb}
     \end{subfigure}
     \hfill
     \begin{subfigure}[b]{.23\textwidth}
         \centering
         \includegraphics[width=\textwidth, height=.75\textwidth]{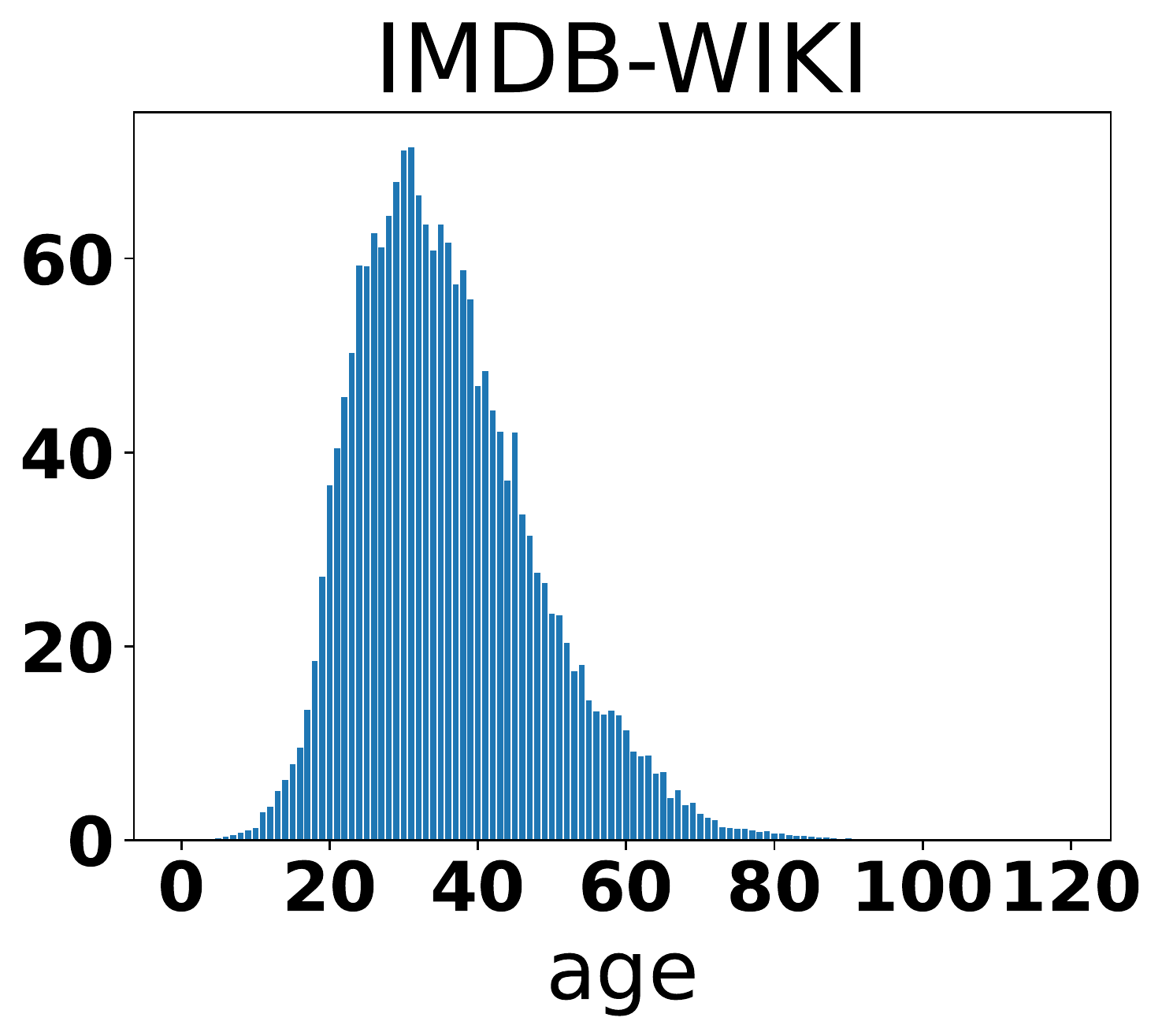}
         \label{fig:hist_imdb}
     \end{subfigure}
     \hfill
     \begin{subfigure}[b]{.23\textwidth}
         \centering
         \includegraphics[width=\textwidth, height=.75\textwidth]{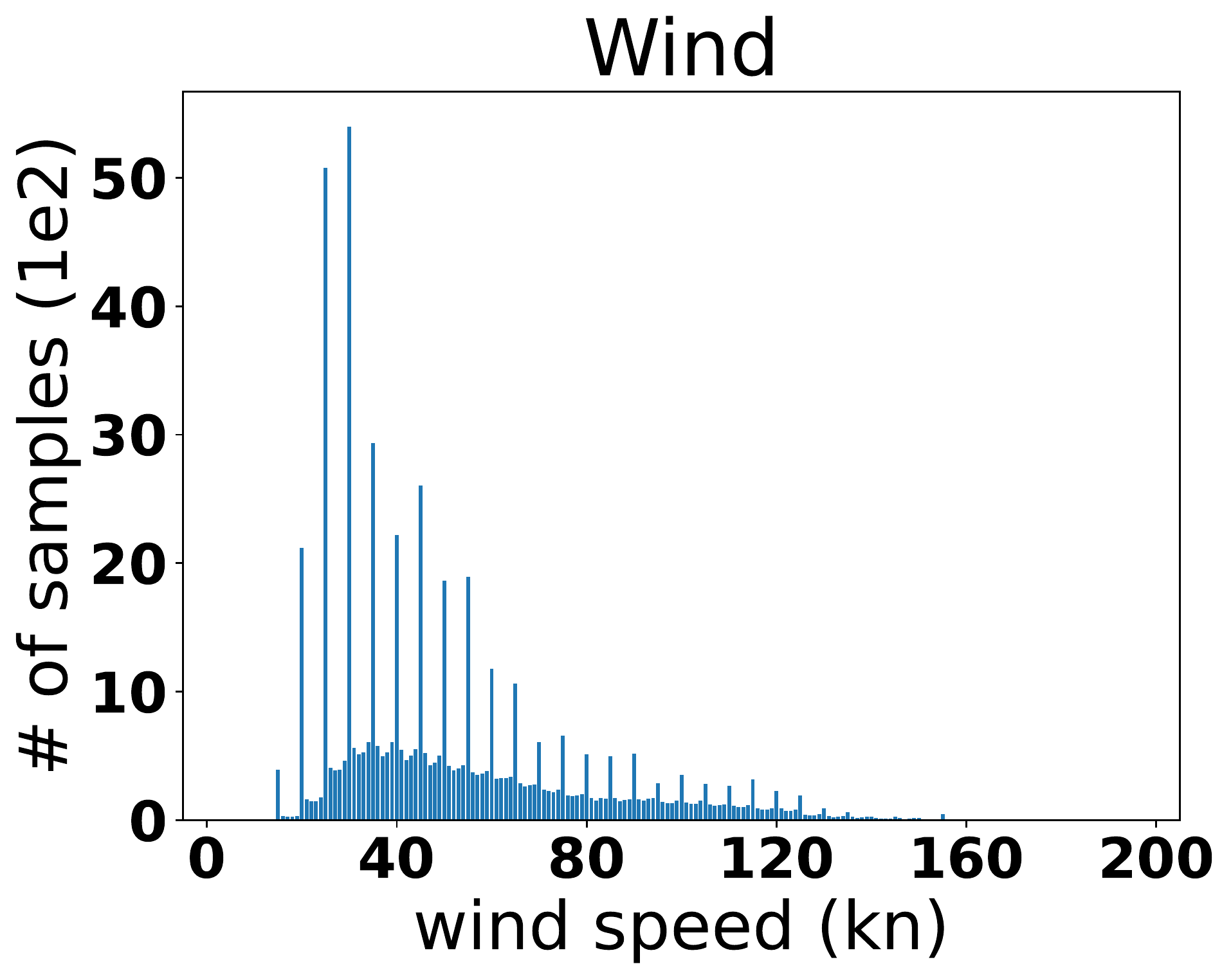}
         \label{fig:hist_wind}
     \end{subfigure}
     \begin{subfigure}[b]{.23\textwidth}
         \centering
         \includegraphics[width=\textwidth, height=.75\textwidth]{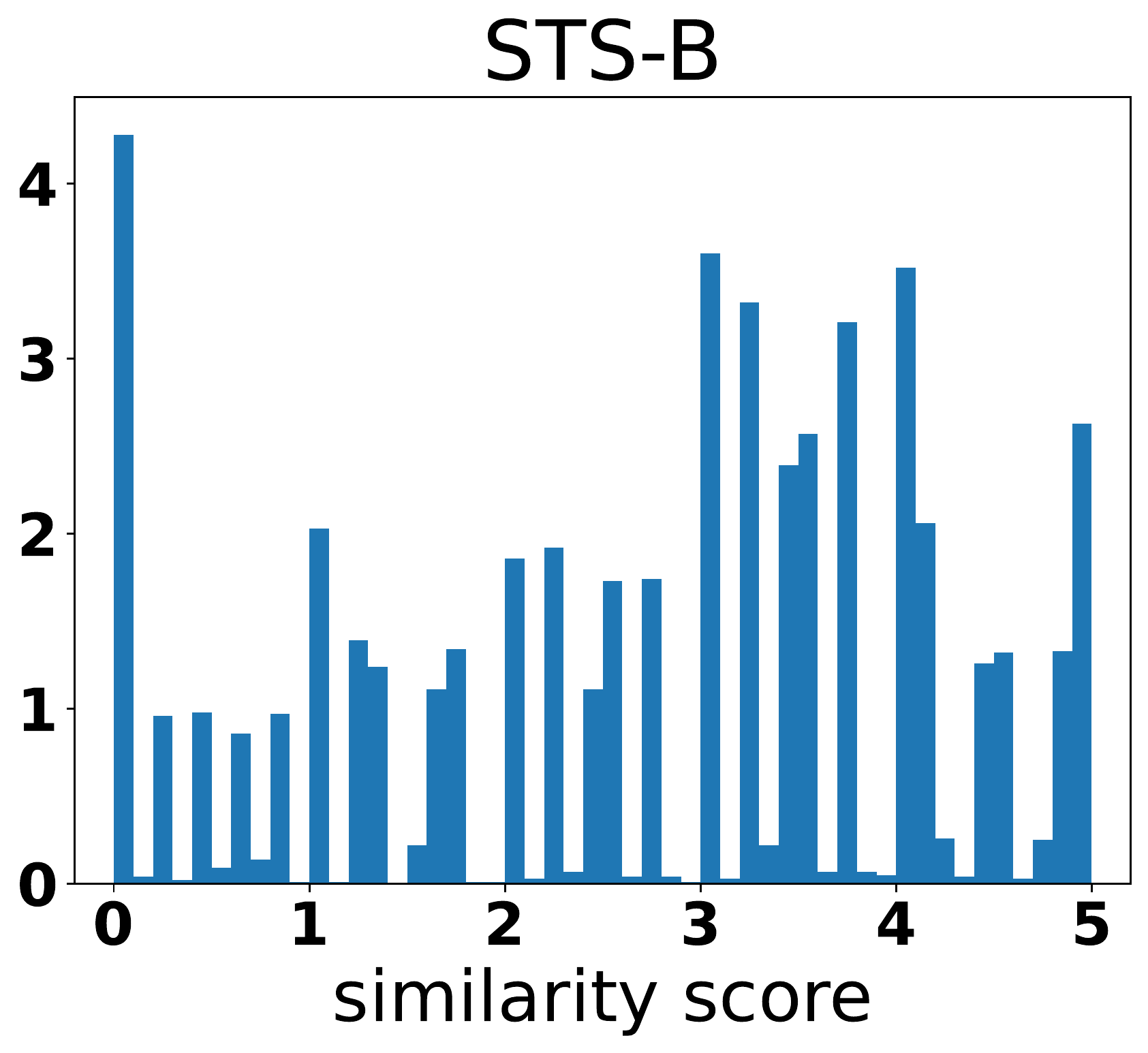}
         \label{fig:hist_sts}
     \end{subfigure}
     \vspace{-20pt}
        \caption{\textbf{Dataset overview.} Distribution of the training set of the four datasets. We consider very different tasks ranging from age regression, to text similarity prediction, and wind speed estimation. }
        \label{fig:dataset_hist}
\end{figure}

Fig. \ref{fig:dataset_hist} shows an overview of the distribution of all datasets. Note the irregular distribution of the Wind dataset, potentially caused by rounding artifacts, but already present in the dataset's release article (see Fig. 2 of \cite{maskey2020Winddeepti}). \textbf{STS-B} also displays an irregular distribution, likely linked to the similarity scores obtained by averaging values from multiple subjective human annotations. We find experimentally that estimating frequencies with kernel density estimation (KDE) leads to more robust performance than with simple histograms for such irregular distributions. We thus replace $f_{b(n)}$ in \cref{equation:weight_fre_bin} with the sample-level estimated density for these two datasets: 
\begin{equation}
\hat{f}(x) = \frac{1}{Nh}\sum_{n=1}^{N}K\left(\frac{x - x_n}{h}\right) \:,
\end{equation}
with $K$ the Gaussian kernel, and $h$ set to $2$ for \textbf{Wind}, and $0.5$ for \textbf{STS-B}. Methods using KDE for frequency estimation are further denoted by $\kappa$ .

\subsection{Experimental Setup}

\paragraph{\bf Competing methods}
We benchmark our approach against a vanilla backbone and a comprehensive set of recent state-of-the-art methods:
\begin{itemize}
    \item \textbf{Vanilla:} baseline without specific technique for imbalanced regression.
    \item \textbf{RRT:} (regressor re-training) two-stage training method: after training the model normally in the first stage, the regressor is re-trained with a rebalanced loss function.
    \item \textbf{SQINV/INV:} inverse frequency weighting (for \textbf{STS-B}) and inverse-squareroot frequency weighting (for the remaining datasets) in the loss function, following~\cite{yang2021delving}.
    \item \textbf{LDS+FDS:} label and feature smoothing \cite{yang2021delving}.
    \item \textbf{RankSim:} regularization technique proposed by \cite{gong2022ranksim}, which encourages data points that have more similar target values to also lie closes to each other in feature space, by aligning the corresponding rank orders.
    \item \textbf{BalancedMSE} introduced by \cite{ren2022balancedMSE} integrates the training label distribution prior into the conventional L2 loss.
    \item \textbf{DenseWeight} \cite{steininger2021denseWeight} is a kernel-based weighting method to enhance traditional reweighting approaches.
\end{itemize}

The baselines methods are to some degree complementary and can be combined. Following \cite{yang2021delving,gong2022ranksim}, we also test combinations of them as. e.g., LDS+FDS is a reweighting of features and loss terms, and RankSim is an additional regularizer that can be combined with different loss functions. For \textbf{AgeDB}, \textbf{IMDB-WIKI}, and \textbf{STS-B} datasets, the performance metrics for the baselines RankSim and LDS+FDS are taken from \cite{yang2021delving,gong2022ranksim}. We also take the performance metrics of BalancedMSE on dataset \textbf{IMDB-WIKI} from \cite{ren2022balancedMSE}. For other experiments, we run the baselines based on their public implementations.
%
%
\paragraph{\bf Metrics} Following \cite{yang2021delving}, we report the Mean Absolute Error (MAE)
on the \textbf{AgeDB}, \textbf{IMDB-WIKI}, and \textbf{Wind} datasets. 
To be comparable, we follow \cite{cer2017semantic_STS_B} for 
\textbf{STS-B} and report the Pearson correlation coefficient, expressed as a percentage ($P\%$). 
%
We report these metrics on the complete test set (All), as well as separately for different data density regimes. To that end the test data are binned into a frequency distribution. Bins with \textgreater100 samples form the many-shot regime ( \emph{many} in the tables), bins with 20 to 100 samples form the medium-shot (\emph{med.}) regime, and bins with \textless20 samples are the few-shot (\emph{few}) regime~\cite{yang2021delving}. Similar to other studies, we use bins of size $1$ on \textbf{AgeDB}, \textbf{IMDB-WIKI}, and \textbf{Wind}, and of size $0.1$ on \textbf{STS-B}.
We report the Uncertainty Calibration Error (UCE) to evaluate the quality of the predicted uncertainties. 

\paragraph{\bf Implementation details} 
The code to conduct the experiments is implemented in Pytorch \cite{paszke2019pytorch}. We use ResNet-50 as backbone for \textbf{AgeDB} and \textbf{IMDB-WIKI},  and ResNet-18 for \textbf{Wind}. For \textbf{STS-B}, we use BiLSTM+GloVe as the baseline, following \cite{wang2018glue_STS_B}. Across datasets, each expert head $G_m$ is implemented with a single linear layer, incurring a marginal parameter cost. The number of experts in \MethodName ($M$) is tuned by training different instances and selecting the best one based on validation set performance. This gives $M=2$ for \textbf{AgeDB} and \textbf{Wind}, and, $M=3$ for \textbf{IMDB-WIKI} and \textbf{STS-B}. For further details about model training, see the supplementary material.

\begin{table*}[t!]
\centering
\caption{\textbf{Main experiment.} We report the regression performance (MAE$\downarrow$ for AgeDB, IMDB-WIKI, Wind datasets, Pearson correlation (\%)$\uparrow$ for STS-B). 
For each column, the best results are in \textbf{bold} and the second best results are \underline{underlined}.}
\label{table:main_experiment}
\resizebox{.8\textwidth}{!}{
\begin{tabular}{lccccccccc}
\toprule
\multicolumn{1}{c}{} & \multicolumn{4}{c}{\textbf{AgeDB} $\downarrow$} &\phantom{a}& \multicolumn{4}{c}{\textbf{IMDB-WIKI} $\downarrow$} \\  \cmidrule{3-4} \cmidrule{8-9}
\multicolumn{1}{c}{}    & \multicolumn{1}{c}{\textbf{All}} & \multicolumn{1}{c}{Many} & \multicolumn{1}{c}{Med.} & \multicolumn{1}{c}{Few} && \multicolumn{1}{c}{\textbf{All}} & \multicolumn{1}{c}{Many} & \multicolumn{1}{c}{Med.} & \multicolumn{1}{c}{Few}\\ \cmidrule{2-5} \cmidrule{7-10}
Vanilla & 7.77 & 6.62 & 9.55 & 13.67            &&        8.06  &   7.23  &   15.12  &   26.33      \\
+RankSim & 7.13 & \underline{6.51} & 8.17 & 10.12           &&                7.72  & 6.93 &  14.48  & 25.38    \\  \cmidrule{2-5} \cmidrule{7-10}
RRT & 7.74 & 6.98 & 8.79 & 11.99                &&        7.81  &  7.07  &  14.06 &   25.13  \\
+LDS,FDS & 7.66 & 6.99 & 8.60 &  11.32          &&        7.65   &  7.06  &   12.41  &   23.51      \\
+RankSim & 7.11 & 6.53 & 8.00 & 10.04           &&    7.55  &   6.83   &  13.47  &   24.72          \\
+LDS,FDS+RankSim & 7.13 & 6.54 & 8.07 & 10.12   &&\underline{7.37}   &  \textbf{6.80}& \underline{11.80} & 23.11            \\  \cmidrule{2-5} \cmidrule{7-10}
SQINV & 7.81 & 7.16 & 8.80 & 11.20              &&  7.87  &  7.24  &  12.44  &  22.76               \\
+LDS,FDS & 7.55 & 7.01 & 8.24 & 10.79           &&     7.78  &  7.20 &   12.61  &  22.19            \\
+RankSim &\underline{6.91} & \textbf{6.34} & 7.79 & \underline{9.89}    &&   7.42  &  6.84  &  12.12  &  22.13              \\
+LDS,FDS+RankSim & 7.03 & 6.54 & \underline{7.68} & 9.92    &&   7.69  &  7.13  &  12.30  &  \underline{21.43}              \\ \cmidrule{2-5} \cmidrule{7-10}

BalancedMSE & 8.02  & 6.78  & 9.98  & 14.30     && 8.08   & 7.52   & 12.47  &  23.29              \\ 
DenseWeight & 8.65 & 8.36 & 8.03 & 13.07  && 7.85  & 7.14  & 13.70 & 25.38 \\ \cmidrule{2-5} \cmidrule{7-10}

\textbf{\MethodName} & \textbf{6.82} & 6.55 & \textbf{7.37} & \textbf{7.80} && \textbf{7.36}  & \underline{6.81} &  \textbf{11.78}  & \textbf{20.96}   \\
\\
\multicolumn{1}{c}{} & \multicolumn{4}{c}{\textbf{Wind} $\downarrow$} &\phantom{a}& \multicolumn{4}{c}{\textbf{STS-B} $\uparrow$} \\  \cmidrule{3-4} \cmidrule{8-9}
\multicolumn{1}{c}{}    & \multicolumn{1}{c}{\textbf{All}} & \multicolumn{1}{c}{Many} & \multicolumn{1}{c}{Med.} & \multicolumn{1}{c}{Few} && \multicolumn{1}{c}{\textbf{All}} & \multicolumn{1}{c}{Many} & \multicolumn{1}{c}{Med.} & \multicolumn{1}{c}{Few} \\ \cmidrule{2-5} \cmidrule{7-10}
Vanilla    &  7.48      & 7.38 &  13.10  & 21.42 &   &74.2& 72.0& 62.7& 75.2  \\
+RankSim   &   \underline{7.43}     &\underline{7.33} & 12.49   & 20.50     &&76.8 &71.0 &\textbf{72.9}& 85.2  \\ \cmidrule{2-5} \cmidrule{7-10}
RRT         &  7.51      &7.39 &  13.67  & 22.79    &&74.5& 72.4& 62.3 &75.4 \\
+LDS,FDS  &    7.52    & 7.40 &  13.64  & 22.35     && 76.0 &73.8 & 65.2 &76.7 \\
+RankSim       &   7.44     & 7.34 & 12.73 & 21.03  && \underline{77.1}& 72.2& 68.3& \textbf{86.1}  \\
+LDS,FDS+RankSim   &  7.45 & 7.35 &  12.75  & 20.93 &&  76.6 &71.7 &68.0& \underline{85.5} \\ \cmidrule{2-5} \cmidrule{7-10}
SQINV/INV    &   7.90     & 7.82 & 11.97  & 20.26     &&72.8 &70.3 &62.5& 73.2 \\
+LDS,FDS  &  7.75      & 7.68 & 11.98   & \underline{15.87}     &&76.0 &\underline{74.0}& 65.2 &76.6 \\
+RankSim   &   7.79     &7.71 &  12.22  & 20.07     &&69.9 &65.2 &60.1& 76.0 \\
+LDS,FDS+RankSim  & 7.71  & 7.63 & 12.16 & 16.70    &&75.8 &70.6& 69.0 &82.7  \\\cmidrule{2-5} \cmidrule{7-10}

BalancedMSE & 7.59  & 7.52  & \underline{11.18}  & 17.80     &&  73.7  & 71.4   & 60.8  & 75.9               \\
DenseWeight & 8.28 & 8.17 & 14.40 &  25.42 && 72.9  & 69.6 & 71.7 & 70.7  \\ \cmidrule{2-5} \cmidrule{7-10}

\textbf{\MethodName} $\kappa$ &   \textbf{7.30}     & \textbf{7.23} & \textbf{11.09}   & \textbf{15.43}   && \textbf{77.7}  & \textbf{74.8}  & \underline{72.0}   & 78.9  \\

\bottomrule

\end{tabular}
}
\end{table*}

\subsection{Imbalanced Regression Experiment}

\paragraph{\bf Comparison to state-of-the-art}
We report the quantitative results of our experiments in Table. \ref{table:main_experiment}. In terms of overall performance, \MethodName outperforms all existing approaches on all four datasets.
On \textbf{AgeDB}, \textbf{IMDB-WIKI}, and \textbf{Wind}, our work also achieves the best performance on the \emph{medium-shot} and \emph{few-shot} regions of the distribution. The gain in few-shot performance compared to the Vanilla model ranges from $43\%$ on \textbf{AgeDB} to $20\%$ on \textbf{IMDB-WIKI}. The margin w.r.t.\ the closest competitor ranges from $21\%$ on \textbf{AgeDB} to $3\%$ on \textbf{Wind}. At the same time, \MethodName reaches the best performance in the data-rich region (\emph{many}) on \textbf{Wind}, \textbf{STS-B} and near-best results on the other two datasets, highlighting that it indeed leverages the predictions of different experts to respond to imbalanced datasets with large density variations.
%
On the \textbf{STS-B} dataset, \MethodName achieves the highest Pearson correlation overall, as well as in the \emph{many-density} regime, and the second-highest one for the \emph{medium-shot} regime.
In the \emph{few-shot} setting, \MethodName outperforms the Vanilla model and baselines like RRT, LDS+FDS, and INV. It does, however, perform inferior compared to Vanilla+RankSim, RRT+RankSim, RRT+LDS,FDS+RankSim, and SQINV/INV+LDS,FDS+RankSim. Nonetheless, \MethodName does not only outperform all competing methods on the entire dataset, but also shows well-balanced performance across all three density parts.
%
We present more experiments with different baselines combined with the light ensembling strategy of \MethodName in the supplementary material. Our method still largely outperforms those models, especially in the \emph{few-shot} setting.

In summary, \MethodName sets a new state of the art for all four datasets.
%
\MethodName is very flexible and can be readily adapted to different tasks and instantiated with different encoder and decoder architectures.
It comes at a marginal computational cost. For instance, when using ResNet50 as backbone, each expert only increases the parameter count by 0.01\%.


\paragraph{\bf Uncertainty prediction} 
We report the uncertainty calibration metrics in Table. \ref{table:UCE-alldatasets}. 
Unsurprisingly, we observe  uncertainty as well is more difficult to estimate in the few-shot regime, i.e., in areas of low sample density. \MethodName outperforms the vanilla network, trained with NLL loss, and the largest gains occur in the few-shot regime. e.g., the uncertainty calibration error (UCE) for samples from few-shot regions drops by 41\% on \textbf{AgeDB} and by 37\% on \textbf{Wind}.

\begin{table}[h!]
\centering
\caption{\textbf{Uncertainty calibration}. UCE of \MethodName vs. NLL. }
 \label{table:UCE-alldatasets}
\resizebox{.5\textwidth}{!}{
\begin{tabular}{lccccccccc}
\toprule
\multicolumn{1}{c}{} & \multicolumn{4}{c}{\textbf{AgeDB} $\downarrow$} &\phantom{a}& \multicolumn{4}{c}{\textbf{IMDB-WIKI} $\downarrow$} \\  \cmidrule{3-4} \cmidrule{8-9}
\multicolumn{1}{c}{}    & \multicolumn{1}{c}{\textbf{All}} & \multicolumn{1}{c}{\textbf{Many}} & \multicolumn{1}{c}{\textbf{Med.}} & \multicolumn{1}{c}{\textbf{Few}} && \multicolumn{1}{c}{\textbf{All}} & \multicolumn{1}{c}{\textbf{Many}} & \multicolumn{1}{c}{\textbf{Med.}} & \multicolumn{1}{c}{\textbf{Few}} \\ \cmidrule{2-5} \cmidrule{7-10}
NLL & 1.76 & 1.05 & \textbf{2.66} & 6.01            &&        2.36  &   1.83  &   7.37  &   17.71      \\
\textbf{\MethodName} & \textbf{1.08} & \textbf{0.72}  & 2.69  & \textbf{3.54} && \textbf{1.94} & \textbf{1.37}  & \textbf{6.57}  & \textbf{15.74} \\ 

\multicolumn{1}{c}{} & \multicolumn{4}{c}{\textbf{Wind} $\downarrow$} &\phantom{a}& \multicolumn{4}{c}{\textbf{STS-B} $\downarrow$} \\  \cmidrule{3-4} \cmidrule{8-9}
\multicolumn{1}{c}{}    & \multicolumn{1}{c}{\textbf{All}} & \multicolumn{1}{c}{\textbf{Many}} & \multicolumn{1}{c}{\textbf{Med.}} & \multicolumn{1}{c}{\textbf{Few}} && \multicolumn{1}{c}{\textbf{All}} & \multicolumn{1}{c}{\textbf{Many}} & \multicolumn{1}{c}{\textbf{Med.}} & \multicolumn{1}{c}{\textbf{Few}} \\ \cmidrule{2-5} \cmidrule{7-10}
NLL & 6.68 & 6.59  & 11.52  & 20.95  && 0.68  & 0.65  & 0.77  & 0.71\\
\textbf{\MethodName} & \textbf{6.49} &\textbf{ 6.44}  & \textbf{9.45}  & \textbf{13.20} && \textbf{0.64} & \textbf{0.63}  & \textbf{0.68}  &\textbf{0.68} \\ 

\bottomrule

\end{tabular}
}
\end{table}


\subsection{Ablation Study}
\label{subsec:ablation}

We investigate the contribution of different design choices in our method by training the following variants on the same benchmark data.
%
    
    \textbf{NLL}: 
    The Vanilla architecture, but trained with negative log-likelihood loss (NLL), instead of a standard L1 or L2 loss.  
    
    \textbf{2-branch}, \textbf{3-branch}: The multi-head setup of our model, with sample weighting and dynamic learning, but without uncertainty estimation, corresponding to a naive model ensemble. We train both a two-branch (2-branch) and a three-branch (3-branch) version.
    
    \textbf{No weighting}: Our method without the sample weighting of Eq. \ref{equation:weight_fre_bin}. In that setting, all experts are trained with the same unweighted NLL loss.
    
    \textbf{No Dynamic Learning (No DyL)}: In that setting, we turn off the dynamic training of Eq. \ref{eq:dyl}, hence all experts are jointly trained from the start. 
    
    \textbf{avg-vote}: This approach only differs from \MethodName in that it combines the expert predictions by averaging, rather than based on the estimated uncertainty.
    
    \textbf{oracle-vote}: As an upper bound for the performance of \MethodName, we also report the performance it would achieve if it had access to an oracle that selects the best expert for each data point (instead of using the predicted uncertainty).






\begin{table*}[t!]
\caption{\textbf{Ablation study}. We report the regression performance (MAE$\downarrow$ for AgeDB, IMDB-WIKI, Wind datasets, Pearson correlation (\%)$\uparrow$ for STS-B) for simplified baseline variants of \MethodName.}
\label{table:ablation-variants-model}
      \centering
      \resizebox{.8\textwidth}{!}{
      
      \begin{tabular}{lccccccccc}
\toprule
\multicolumn{1}{c}{} & \multicolumn{4}{c}{\textbf{AgeDB} $\downarrow$} &\phantom{a}& \multicolumn{4}{c}{\textbf{IMDB-WIKI} $\downarrow$} \\  \cmidrule{3-4} \cmidrule{8-9}
\multicolumn{1}{c}{}    & \multicolumn{1}{c}{\textbf{All}} & \multicolumn{1}{c}{Many} & \multicolumn{1}{c}{Med.} & \multicolumn{1}{c}{Few} && \multicolumn{1}{c}{\textbf{All}} & \multicolumn{1}{c}{Many} & \multicolumn{1}{c}{Med.} & \multicolumn{1}{c}{Few} \\ \cmidrule{2-5} \cmidrule{7-10}
Vanilla & 7.77 & 6.62 & 9.55 & 13.67            &&        8.06  &   7.23  &   15.12  &   26.33      \\
\cmidrule{2-5} \cmidrule{7-10}
NLL  & 7.05 & \textbf{6.24} & 8.11 & 11.80    &&  7.57 & \underline{6.81} & 13.95 & 25.67  \\ 
2-branch      & 7.68 & 7.03 & 8.80 & 10.66      &&       7.86 & 7.27 &  12.69  & 22.70\\
3-branch   & 7.80 & 7.19 & 8.93 & 10.44      &&       7.61 & 7.03 &  12.21  & 22.46  \\

No weighting & 7.25  & 6.46 &8.39  &11.56       &&   7.64     & 6.86 & 14.22   & 25.30  \\
No DyL  & 7.60  & 7.43  & 7.82  & \underline{8.62}      &&   8.13     & 7.59 & 12.40   & 21.99  \\

avg-vote      & \textbf{6.81} & \underline{6.36} & \underline{7.61} & 8.89       &&       \textbf{7.34}  & \textbf{6.77} &  \underline{12.06}  & \underline{21.30} \\
oracle-vote      & \color{blue}6.13 & \color{blue}5.76 & \color{blue}6.85 & \color{blue}7.59       &&       \color{blue}6.86     & \color{blue}6.31 &  \color{blue}11.29  & \color{blue}20.59  \\
\cmidrule{2-5} \cmidrule{7-10}
\textbf{\MethodName}  & \underline{6.82} & 6.55 & \textbf{7.37} & \textbf{7.80} &&       \underline{7.36}  & \underline{6.81} &  \textbf{11.78}  & \textbf{20.96}   \\
\\
\multicolumn{1}{c}{} & \multicolumn{4}{c}{\textbf{Wind} $\downarrow$} &\phantom{a}& \multicolumn{4}{c}{\textbf{STS-B} $\uparrow$} \\  \cmidrule{3-4} \cmidrule{8-9}
\multicolumn{1}{c}{}    & \multicolumn{1}{c}{\textbf{All}} & \multicolumn{1}{c}{Many} & \multicolumn{1}{c}{Med.} & \multicolumn{1}{c}{Few} && \multicolumn{1}{c}{\textbf{All}} & \multicolumn{1}{c}{Many} & \multicolumn{1}{c}{Med.} & \multicolumn{1}{c}{Few} \\ \cmidrule{2-5} \cmidrule{7-10}
Vanilla    &  7.48      & 7.38 &  13.10  & 21.42 &   &74.2& 72.0& 62.7& 75.2  \\
\cmidrule{2-5} \cmidrule{7-10}
NLL   &    \underline{7.36}    & 7.26 & 12.74   & 22.67      && 76.2   &73.5   & 68.7   & 76.8 \\ 
2-branch $\kappa$        & 7.56       & 7.49 & \underline{11.15}   & 17.83   && 76.0  & 73.4  & 68.2   & 74.0  \\
3-branch  $\kappa$ &  7.64      & 7.54& 12.76   & 19.63  && 75.9  & 72.5  & \underline{71.8}   & 74.5  \\

No weighting  & 7.36  & 7.25  & 12.63 & 22.98      &&   75.3     & 72.2  & 68.0   & 76.0  \\
No DyL $\kappa$  & 8.13 & 8.06 & 12.10 & \textbf{15.39}     &&    75.4    & 72.0  & 69.2   & \underline{78.5}   \\

avg-vote  $\kappa$      &   \textbf{7.30}      & \textbf{7.23} & 11.18   & 15.90  && \underline{77.6}  & \underline{74.7}  & \underline{71.8}   & \textbf{78.9}  \\
oracle-vote  $\kappa$     & \color{blue}7.22      & \color{blue}7.15 &  \color{blue}10.91  &  \color{blue}15.33     &&      \color{blue}77.9   & \color{blue}75.2  & \color{blue}72.1   & \color{blue}78.9   \\
\cmidrule{2-5} \cmidrule{7-10}
\textbf{\MethodName  $\kappa$ }   &  \textbf{7.30}     & \textbf{7.23} &  \textbf{11.09}   & \underline{15.43}  &&  \textbf{77.7}  & \textbf{74.8}  & \textbf{72.0}   & \textbf{78.9} \\
\bottomrule
        \end{tabular}
}
        
\end{table*}

\paragraph{\bf Probabilistic training} Training a single-head regression network with the NLL loss instead of the standard regression loss (Vanilla)
already leads to an increase in overall performance across datasets. On three of the four datasets, performance in the \emph{few-shot} regime also improves by $1-2$pts.  In other words, a probabilistic training objective by itself already mitigates the imbalance problem to some degree. This is also clearly visible when comparing the performance of the 2/3-branch model and the avg-vote variant of our method. Both aggregate the experts' prediction by averaging, and only differ in the applied loss. The avg-vote model, trained with NLL, outperforms the 2/3-branch model on all parts of the distribution across all datasets. Our results support the practice of replacing a standard regression loss with NLL for imbalanced regression problems.

\paragraph{\bf Uncertainty voting} In addition to improving the overall performance, the NLL training objective we use in \MethodName allows us to select the best prediction based on aleatoric uncertainty. Compared to the 2/3-branch models, this brings a more significant improvement in the \emph{few-shot} regime, without sacrificing performance in \emph{many-shot} regions. For instance, \MethodName reduces the error on \textbf{AgeDB}, by $5.87$pt in the \emph{few-shot} region, compared to only $1.87$pt and $3.23$pt reductions with the NLL and 3-branch models, respectively.  At the same time, \MethodName  outperforms the Vanilla model in the \emph{many-shot} regime. This highlights how uncertainty-based voting helps to select the correct expert at inference time, which also becomes apparent when comparing \MethodName against the avg-vote model. While overall performance is similar, uncertainty voting excels in the \emph{few-shot} regions and consistently beats average voting. We emphasize that once the mixture of expert has been trained with NLL, uncertainty voting comes at \emph{no-cost} compared to traditional average ensembling. Average voting only seems beneficial in the \emph{many-shot} regions of \textbf{AgeDB} and \textbf{IMDB-WIKI}, where it brings the benefit of a traditional model ensemble.  
Lastly,  a comparison of \MethodName with oracle-voting bound demonstrates that the proposed uncertainty-based expert selection achieves near-perfect decisions in the medium- and low-density regions. The role of the uncertainty-based selection is to close the gap in performance between the naive ensembling (avg-vote) and the oracle-vote. Expressing the performance improvement achieved by uncertainty-voting in terms of the percentage of that gap, we obtain $32\%$ and $84\%$ in the Med. and Few regions of AgeDB for example. 
While better uncertainty calibration would certainly help to further improve the results on \textbf{AgeDB} and \textbf{IMDB-WIKI}, it seems that the per-expert predictions, rather than the voting mechanism, are the current bottleneck for \textbf{Wind} and \textbf{STS-B}.

\paragraph{\bf Multi-head structure} 
A multi-head architecture with specialized heads also generally boosts overall performance, even without the probabilistic loss. The two and three-branch models (2-branch and 3-branch) improve performance primarily in the \emph{few-shot} regions of the distribution, by $\approx3$pt MAE on \textbf{AgeDB}, \textbf{IMDB-WIKI}, and \textbf{Wind}, and by $\approx1$pt $P\%$ on \textbf{STS-B}. However, these models tend to suffer in high-density regions: \emph{many-shot} performance is lower on three datasets for the 2-branch model and on two datasets for the 3-branch model.

\section{Conclusion}
We have proposed \MethodName, a simple and effective method for deep imbalanced regression. Our method, which can be understood as an ensemble over variably rebalanced regressors with uncertainty-based aggregation, can be freely combined with different encoder backbones and comes with negligible computational overhead. 
Integrating ensemble learning with uncertainty estimation enables to design a dynamic voting system in a single end-to-end training. 
In experiments on four different datasets, \MethodName reaches the best overall performance and sets a new state of the art. Importantly, our method decreases the prediction error particularly in under-represented, low-density regions, while maintaining excellent performance in high-density regions. By construction, \MethodName provides well-calibrated predictive uncertainties along with the target values, which enhance interpretability of the results and aid downstream tasks.

%
%
%
%

\bibliographystyle{splncs04}
\bibliography{main}

\clearpage
\setcounter{page}{1}
\textbf{\huge{Supplementary Material}}


\section{Experimental Settings}

\subsection{Training Details}
\paragraph{AgeDB}
 We use a ResNet-50 backbone for all methods and train each model for 90 epochs with a batch size of 64 and Adam optimizer. The initial learning rate is set as $1 \times 10^{-3}$ and is scheduled to drop by a factor 10  at epochs 60 and 80. For the last output layers of uncertainty estimation, we set a smaller learning rate, $1 \times 10^{-4}$ for stable training. In the second training stage of RRT, we use an initial learning rate of $1 \times 10^{-4}$ and train the model for a total of 30 epochs. We use $L_1$ loss for baselines and Laplacian negative log-likelihood loss for our proposed method. For kernel density estimation, we use the Gaussian kernel with bandwidth 2.

\paragraph{IMDB-WIKI}
We use ResNet-50 for all experiments and train each model for 90 epochs with batch size 256 and Adam optimizer. The initial learning rate is set as $1 \times 10^{-3}$ and it is divided by 10 at epochs 60 and 80. For last output layers of uncertainty estimation, we set a smaller learning rate, $1 \times 10^{-4}$ for stable training. During the second training stage of RRT, we set the initial learning rate to $1 \times 10^{-4}$ and conducted training for a total of 30 epochs. We use $L_1$ loss for baselines and Laplacian negative log-likelihood loss for our proposed method. For kernel density estimation, we use the Gaussian kernel with bandwidth 2. 

\paragraph{Wind}
We use ResNet-18 for all experiments and train each model 90 epochs with batch size 64 and Adam optimizer. The initial learning rate is set as $1 \times 10^{-3}$ and it is scheduled to drop by 10 times at epoch 60 and 80. For last output layers of uncertainty estimation, we set a smaller learning rate, $1 \times 10^{-4}$ for stable training. In the second training stage of RRT, we conducted training for a total of 30 epochs with an initial learning rate of $1 \times 10^{-4}$. We use $L_1$ loss for baselines and Laplacian negative log-likelihood loss for our proposed method. For kernel density estimation, we use the Gaussian kernel with bandwidth 2. 

\paragraph{STS-B}
We use a two-layer BiLSTM as the encoder to learn features and then a final regressor to output final predictions. We train each model 200 epochs with batch size 16 and Adam optimizer. The learning rate is $2.5 \times 10^{-4}$. The hyper-parameter settings for RRT remain consistent throughout both the first and second training stages. We use $L_2$ loss for baselines and Laplacian negative log-likelihood loss for our proposed method. For kernel density estimation, we use the Gaussian kernel with bandwidth 0.5. 



\subsection{Evaluation Metrics}

We provide the details of the evaluation metrics:
\begin{itemize}
    \item \textbf{MAE}: the mean absolute error 
    \begin{equation}
            \label{eq:mae}
            MAE = \frac{1}{N} \sum^N_{n=1} |y_n - \hat{y}_n|
        \end{equation}

    
    \item \textbf{RMSE}: the root mean squared error 
    \begin{equation}
            \label{eq:rmse}
            RMSE = \sqrt{\frac{1}{N}\sum^N_{n=1} (y_n - \hat{y}_n)^2}
        \end{equation}

        
    \item \textbf{Pearson correlation}: to evaluate performance on the STS-B dataset
    \begin{equation}
        P\% = \frac{\sum^N_{n=1}(y_n - \bar{y})(\hat{y}_n-\bar{\hat{y}})}{\sqrt{\sum^N_{n=1}(y_n - \bar{y})^2 \sum^N_{n=1}(\hat{y}_n-\bar{\hat{y}})^2}
            } \times 100
        \end{equation}
    \item \textbf{UCE}: the Uncertainty Calibration Error is used to evaluate the quality of the predicted uncertainties
    \begin{equation}
    \label{eq:pearson}
                UCE = \sum^B_{b=1} \frac{N_b}{N} |MAE(b) - \overline{std(b)}|
    \end{equation}
    where $N$ is the total number of samples, $B$ is the total number of bins, $N_b$ is the number of samples falling into bin $b$, $MAE(b)$ is the MAE of samples in bin $b$, $\overline{std(b)}$ is the mean standard deviation of samples in bin $b$, and $\bar{\hat{y}}$  (resp $\bar{y}$) the mean predicted (resp. target) value on the test set.
\end{itemize}

\subsection{Additional Results}

We present additional results on the four datasets in Table \ref{tab:sub_ensemblex_experiment}. For completeness, we report the performance of the combination of  different baselines and the ensembling strategy. In general, we observe that the ensembling strategy results in improvement of the baselines in most cases, with varying impacts across different shot regions. These improvements however do not match the performance of \MethodName, especially in the few-shot region. 

\begin{table*}[t!]
\centering
\caption{\textbf{Additional results with ensembling strategy} We report the regression performance (MAE$\downarrow$ for AgeDB, IMDB-WIKI, Wind datasets, Pearson correlation (\%)$\uparrow$ for STS-B). 
For each column, the best results are in \textbf{bold} and the second best results are \underline{underlined}.}
\label{tab:sub_ensemblex_experiment}
\begin{tabular}{lccccccccc}
\toprule
\multicolumn{1}{c}{} & \multicolumn{4}{c}{\textbf{AgeDB} $\downarrow$} &\phantom{a}& \multicolumn{4}{c}{\textbf{IMDB-WIKI} $\downarrow$} \\  \cmidrule{3-4} \cmidrule{8-9}
\multicolumn{1}{c}{}    & \multicolumn{1}{c}{\textbf{All}} & \multicolumn{1}{c}{Many} & \multicolumn{1}{c}{Med.} & \multicolumn{1}{c}{Few} && \multicolumn{1}{c}{\textbf{All}} & \multicolumn{1}{c}{Many} & \multicolumn{1}{c}{Med.} & \multicolumn{1}{c}{Few}\\ \cmidrule{2-5} \cmidrule{7-10}
Vanilla & 7.77 & 6.62 & 9.55 & 13.67            &&        8.06  &   7.23  &   15.12  &   26.33      \\
+RankSim & \underline{7.13} & \textbf{6.51} & 8.17 & \underline{10.12}           &&                7.72  & 6.93 &  14.48  & 25.38    \\ 
+LDS,FDS & 7.55 & 7.01 & 8.24 & 10.79 && 7.78 & 7.20 & 12.61 & \underline{22.19}\\
BalancedMSE & 8.02 & 6.78 & 9.98 & 14.30 && 8.08 & 7.52 & \underline{12.47} & 23.29 \\
DenseWeight & 8.65 & 8.36 & \underline{8.03} & 13.07  && 7.85  & 7.14  & 13.70 & 25.38 \\ 
\cmidrule{2-5} \cmidrule{7-10}

Ensemble+RankSim & 7.48 & 6.48 & 8.90 & 12.95 && 7.70 & 6.92 & 14.35 & \underline{25.06} \\
Ensemble+LDS,FDS & 7.97 & 7.23 & 9.01 & 12.02 && 8.27 & 7.61 & 13.82 & 23.31 \\
Ensemble+BalancedMSE &7.91  & 7.23  & 8.41 & 13.06 && 7.94 & 7.26 & 13.50 & 25.26 \\
Ensemble+DenseWeight & 7.76 & 7.11 &  8.63 & 11.53 && \underline{7.60} & \underline{6.82} & 14.17 & 25.64 \\
\cmidrule{2-5} \cmidrule{7-10}

\textbf{\MethodName} & \textbf{6.82} & \underline{6.55} & \textbf{7.37} & \textbf{7.80} && \textbf{7.36}  & \textbf{6.81} &  \textbf{11.78}  & \textbf{20.96}   \\
\\
\multicolumn{1}{c}{} & \multicolumn{4}{c}{\textbf{Wind} $\downarrow$} &\phantom{a}& \multicolumn{4}{c}{\textbf{STS-B} $\uparrow$} \\  \cmidrule{3-4} \cmidrule{8-9}
\multicolumn{1}{c}{}    & \multicolumn{1}{c}{\textbf{All}} & \multicolumn{1}{c}{Many} & \multicolumn{1}{c}{Med.} & \multicolumn{1}{c}{Few} && \multicolumn{1}{c}{\textbf{All}} & \multicolumn{1}{c}{Many} & \multicolumn{1}{c}{Med.} & \multicolumn{1}{c}{Few} \\ \cmidrule{2-5} \cmidrule{7-10}
Vanilla    &  7.48      & 7.38 &  13.10  & 21.42 &   &74.2& 72.0& 62.7& 75.2  \\
+RankSim   &   7.43     & 7.33 & 12.49   & 20.50     && \underline{76.8} &71.0 &\textbf{72.9}& \textbf{85.2}  \\ 
+LDS,FDS & 7.75 & 7.68 & 11.98 & \underline{15.87} && 76.0 & \underline{74.0} & 65.2 & 76.6 \\
BalancedMSE & 7.59 & 7.52 & \underline{11.18} & 17.80 && 73.7 & 71.4 & 60.8 & 75.9 \\
DenseWeight & 8.28 & 8.17 & 14.40 &  25.42 && 72.9  & 69.6 & 71.7 & 70.7  \\ \cmidrule{2-5} \cmidrule{7-10}

Ensemble + RankSim & \underline{7.36} & \underline{7.25} & 12.68 & 22.18 && 76.1 &73.4 & 70.3 & 75.7 \\
Ensemble+LDS,FDS &7.78  & 7.69 & 12.56& 18.44 && 75.5 & 71.9 & 68.9 & 76.9 \\
Ensemble+BalancedMSE & 7.71 & 7.63  & 11.51 & 18.55 && 67.8 & 63.6 & 64.0& 64.4\\
Ensemble + DenseWeight & 7.65 & 7.53 & 13.80 & 25.31 && 75.9 & 73.1 & 69.9 & 73.1 \\ \cmidrule{2-5} \cmidrule{7-10}

\textbf{\MethodName} $\kappa$ &   \textbf{7.30}     & \textbf{7.23} & \textbf{11.09}   & \textbf{15.43}   && \textbf{77.7}  & \textbf{74.8}  & \underline{72.0}   & \underline{78.9}  \\

\bottomrule

\end{tabular}
\end{table*}

\begin{figure}[t!]
  \begin{center}
  \includegraphics[width=.4\textwidth]{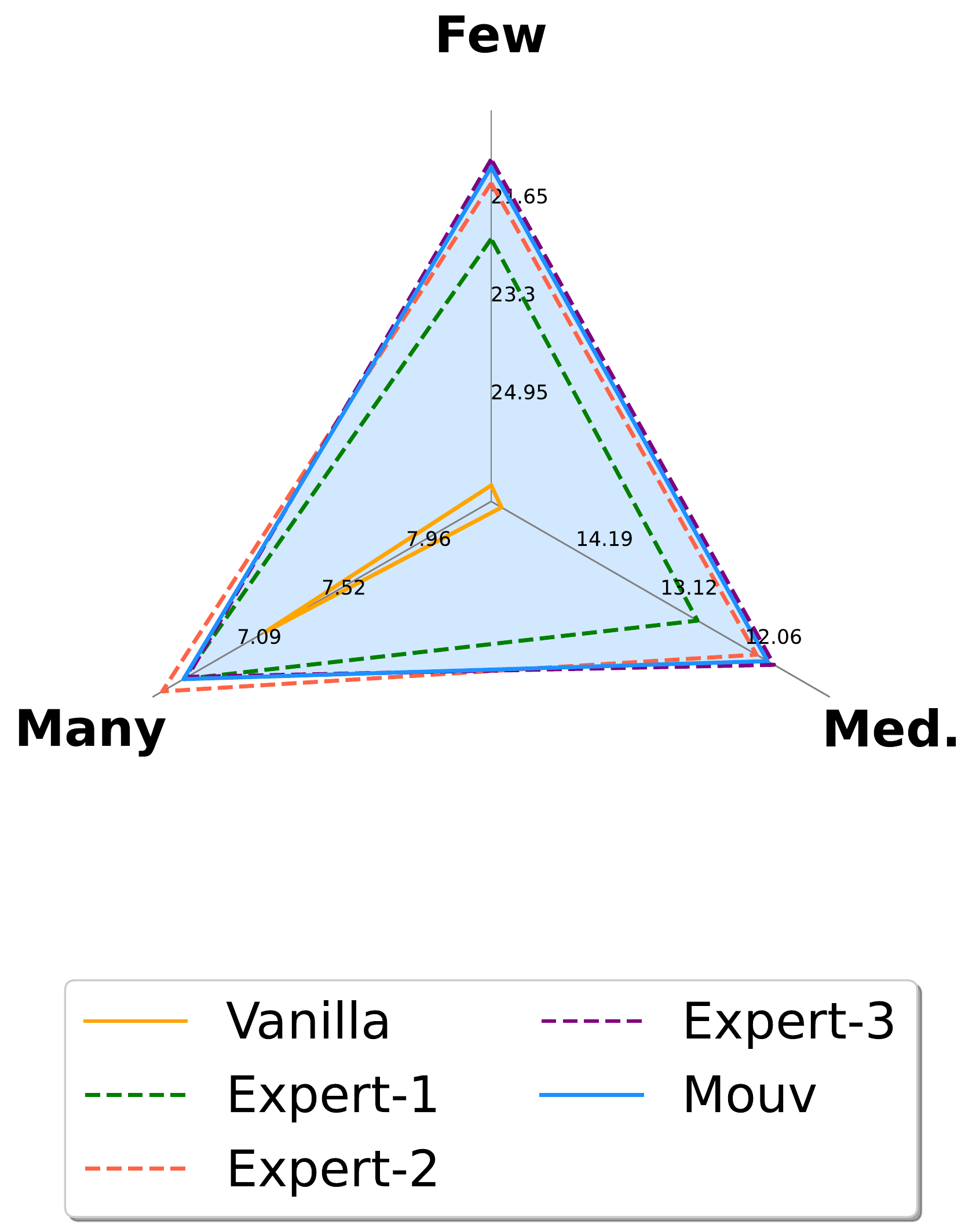}
  \vspace{-10pt}
  \end{center}
    \caption{Per-expert and aggregated MAE on \textbf{IMDB-WIKI}. Larger coverage of the triangle indicates better performance. The uncertainty-based aggregation of \MethodName nearly matches the performance of the best expert on each subset of the test data. }
  \label{fig:result_per_expert_IMDB}
\end{figure}

\begin{figure}[t!]
  \centering
  \includegraphics[trim=0cm 1.4cm 0cm 1.4cm, clip, width=.9\linewidth]{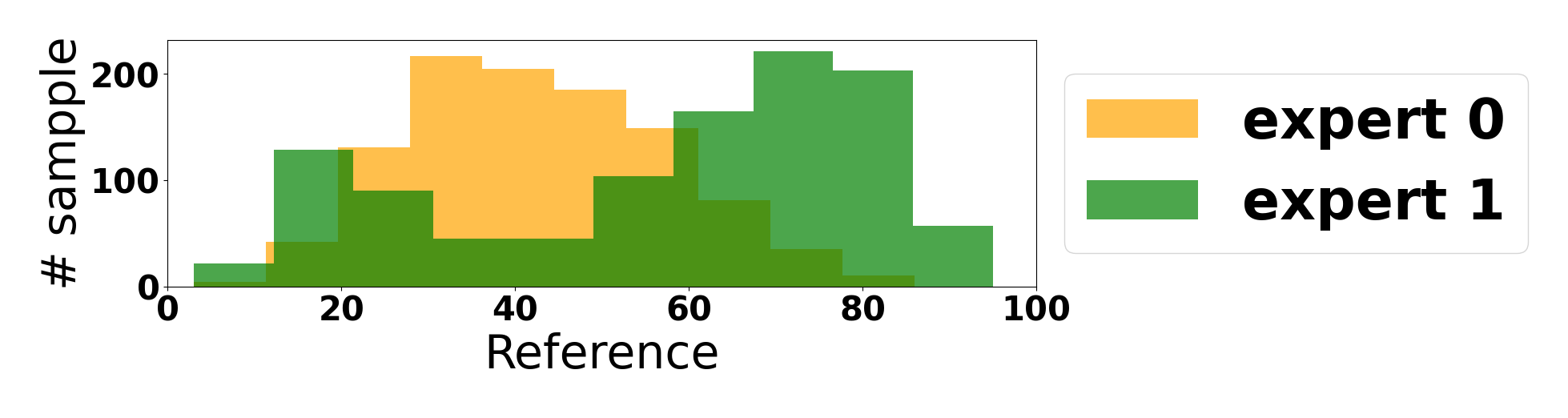}
   \caption{Distribution of expert selection on \textbf{AgeDB}. Notably, Expert 1, which specializes in minority samples, is predominantly chosen for labels in the few-shot regions.
   }
   \label{fig:AgeDB_hist_exp}
\end{figure}

\begin{figure}[t!]
  \centering
  \includegraphics[trim=0cm 1.1cm 0cm 1cm, clip, width=0.8\linewidth]{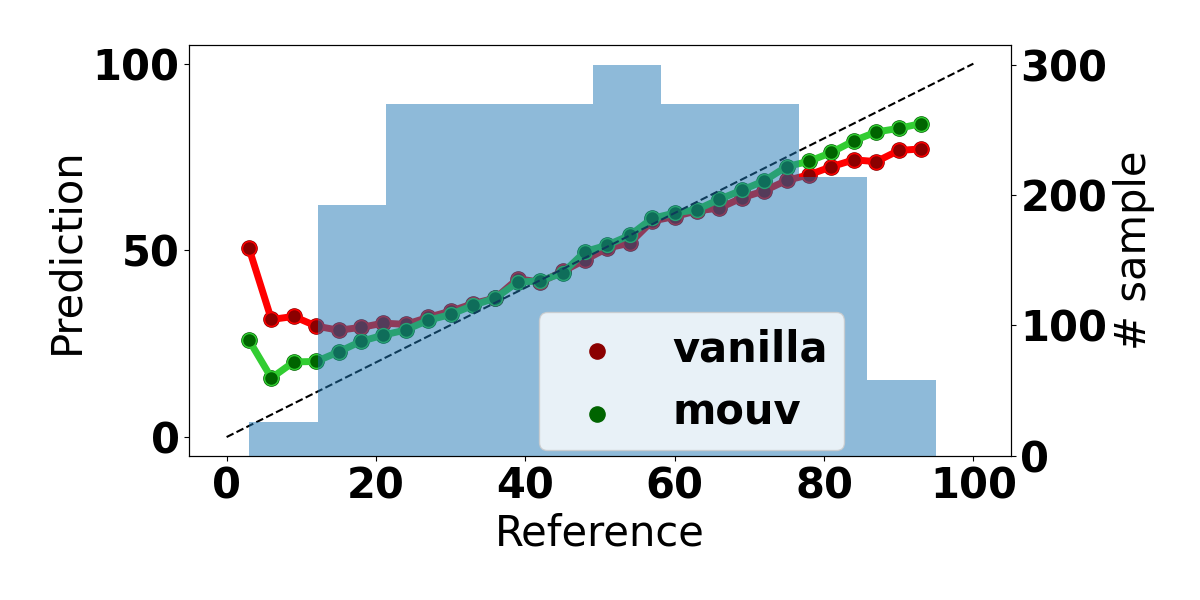}
  \caption{Predictions of ours vs. vanilla model on \textbf{AgeDB}. For clarity we plot the average prediction of each bin. }
   \label{fig:AgeDB_ref_pred}
\end{figure}

We analyze the per-expert and aggregated results in Fig. \ref{fig:result_per_expert_IMDB}, demonstrating that \MethodName's uncertainty voting effectively selects the expert with the best prediction. The ensemble's prediction quality closely matches that of an oracle that always picks the right expert across all data regimes. Additionally, we examine the "workload" distribution between the two experts on AgeDB in Fig.\ref{fig:AgeDB_hist_exp}, showing how experts are chosen during inference. Expert 0, focusing on the majority (middle-age), is predominantly chosen for majority data, while Expert 1, focusing on the tails, is selected for minorities, infants, and seniors. We also compare the results of our method with the vanilla model on AgeDB in Fig. \ref{fig:AgeDB_ref_pred}. The figure clearly shows that our method outperforms the vanilla model, particularly in the tails.

\subsection{Ablation summary}
We present a synthesized version of our ablation results in Table. \ref{table:ablation-percent-AgeDB}, where we show the impact of removing one of the components of \MethodName on the test MAE of AgeDB. Each ablation results in a decrease in performance. Some components such as dynamic learning and probabilistic training have a beneficial effect across the data distribution. The other components are geared towards particularly improving the performance on the \emph{few-shot} region. In that region, the multi-head structure, the sample weighting, and the probabilistic training combined with uncertainty voting all incur a $\sim 50\%$ drop of performance if removed, demonstrating their equally important roles.

\begin{table*}[t!]
\caption{\textbf{Percentage change on AgeDB when switching off the different components of MOUV.} $+$ indicates the increase of MAE$\downarrow$ when the component is switched off.}

    \centering
    \resizebox{\textwidth}{!}{
    
        \begin{tabular}{lccccc|cccc}
             & NLL & MoE & Dyn & min$\sigma$-vote & Weight & All & Many  & Med.  & Few \\ \midrule
        No \textbf{multi-head (MoE)}  & \cmark & \xmark & \xmark & \xmark & \xmark  & +3\% & -5\% & +11\% & +51\% \\
        No \textbf{weighting (Weight)}   & \cmark & \cmark & \cmark & \cmark & \xmark  & +6\% & -1\% & +14\% & +48\% \\
        No \textbf{probabilistic training (NLL)} & \xmark & \cmark & \cmark & \xmark & \cmark  & +13\% & +7\% & +19\% & +37\%\\
        No \textbf{uncertainty voting (min$\sigma$-vote)} & \cmark & \cmark & \cmark & \xmark & \cmark  & 0\% & -3\%&+3\% &+14\% \\
        No \textbf{dynamic learning (Dyn)} & \cmark & \cmark & \xmark & \cmark & \cmark  & +11\% & +13\% & +6\% & +11\%\\
        \end{tabular}
    }
    \label{table:ablation-percent-AgeDB}
\end{table*}

\subsection{Further Analysis}

\paragraph{Repetition Analysis}
We conduct five runs of our method and SQINV+LDS,FDS on the AgeDB and Wind datasets, each  with a different initialization, to assess the impact of randomness in our experiments. Figure \ref{ablation_randomseed_MAE} presents the MAE obtained by \MethodName and SQINV+LDS,FDS on \textbf{AgeDB} and \textbf{Wind} datasets, accompanied by error bars representing the standard deviation across five experiments. In both datasets, we consistently observe that our method achieves smaller mean MAE values across all regions. Specifically, in the \textbf{AgeDB} dataset, the error bars of MAE are relatively small. In the \textbf{Wind} dataset, the error bars of MAE are larger in both methods, particularly in the few-shot region, suggesting more variability in performance.

\begin{figure}[h]
        \begin{subfigure}[b]{\linewidth}
         \centering
        \includegraphics[clip,width=.7\columnwidth]{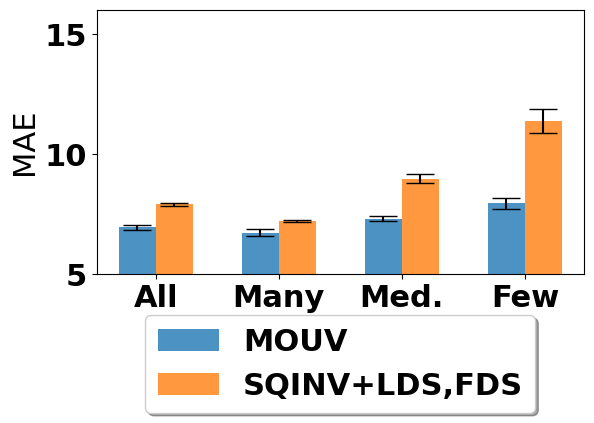}%
        \subcaption{AgeDB}
        \label{fig:ablation_repetition_agedb}
     \end{subfigure}
\end{figure}

\begin{figure}[h]\ContinuedFloat
         \begin{subfigure}[b]{\linewidth}
         \centering
         \includegraphics[clip,width=.7\columnwidth]{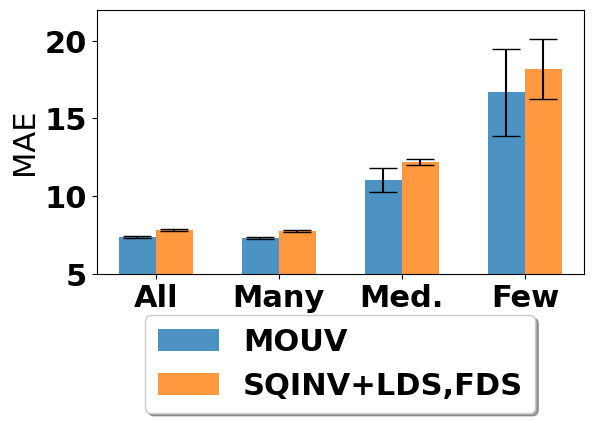}%
         \subcaption{Wind}
        \label{fig:ablation_repetition_wind}
     \end{subfigure}
\caption{MAE of our methods and SQINV+LDS,FDS on AgeDB and Wind datasets obtained across five models trained with different random initialization. The error bar represents the standard deviation of MAEs across various initializations.}
\label{ablation_randomseed_MAE}
\end{figure}

\begin{figure}[t!]
  \centering
    \subfloat[AgeDB]{%
        \includegraphics[clip,width=.8\columnwidth]{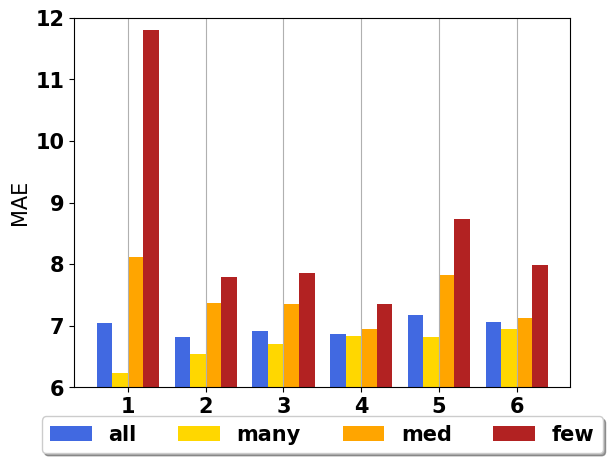}%
        \label{fig:ablation_numbranch_agedb}
    }

    \subfloat[Wind]{%
        \includegraphics[clip,width=.8\columnwidth]{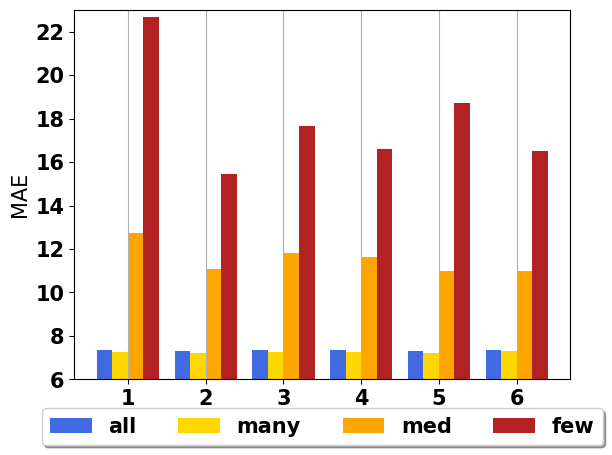}%
        \label{fig:ablation_numbranch_wind}
    }

\caption{MAE for our methods with different number of experts. In both datasets, the MAE drops sharply when transitioning from a model with a single expert to a model with two experts. Adding more experts brings no major improvements when there are already two or three experts.}
\label{ablation_branchnumber_MAE}
\end{figure}

\paragraph{Number of branches.}
The number of branches is the main hyper-parameter of our method. We present the MAE of models with 
$M = {1, 2, 3, 4, 5, 6}$ in Figure \ref{ablation_branchnumber_MAE}. We observe a substantial improvement in MAE within the \emph{few-shot} and \emph{medium-shot} regions when transitioning from a single-expert model to a two-expert model. We note that the performance in the data-scarce parts of the distribution can be further improved by increasing the number of experts, e.g., $M=4$  in \textbf{AgeDB}, which comes, however, at the expense of a slight drop in performance in the many-shot region. In general, our experiments indicate that the configurations with two or three experts achieve best performance across the entire datasets. The optimal number of experts is, of course, problem and dataset-dependent and should be tuned for each application individually. 

\paragraph{Kernel density estimation.}
As a last ablation we employ KDE instead of histogram binning to estimate the sample density. Table \ref{table:ablation_KDE} compares the performance of our method with KDE to the one with simple binning, across all datasets. On \textbf{AgeDB} and \textbf{IMDB-WIKI}, the MAE differences are marginal ($< 0.5$pt) while they are more noticeable on datasets with irregular distribution, especially in the \emph{few-shot} regions. In particular, the MAE drops by 3.61pt in \emph{few-shot} regions of the \textbf{Wind} dataset when using KDE in \MethodName. The result suggests that the more sophisticated approach to density estmation benefits datasets with irregular distributions, while making little difference for datasets with already relatively smooth distributions. We also re-trained the SQINV and RRT methods on \textbf{Wind} with KDE, but did not observe any performance improvement. 

\begin{table}[t!]
  \centering\caption{\textbf{Impact of KDE}. Relative performance variation when switching from histogram-based frequencies to KDE with \MethodName. 
  \underline{Improvements} with KDE are underlined. }
  \label{table:ablation_KDE}
        \begin{tabular}{lcccc}
        \toprule
         &\textbf{ All} & \textbf{Many}& \textbf{Med.} &\textbf{Few} \\  \hline
         
        \textbf{AgeDB $\downarrow$} & +0.21  & +0.43  & \underline{-0.37}  & \underline{-0.24} \\
        \textbf{IMDB-WIKI $\downarrow$} & +0.03 & +0.07 & \underline{-0.37}  & 0\\
        \textbf{Wind  $\downarrow$} & 0 & +0.01  & \underline{-0.45}  & \underline{-3.61}\\ 
        \textbf{STS-B $\uparrow$} & \underline{+1.2} & \underline{+1.9} & -2.5 &  +3.7\\  
        \bottomrule
        \end{tabular}
  \end{table}

\subsection{Dense Regression}

For completeness, we also present results of \MethodName on the structured regression problem of NYUd2 \cite{silberman2012nyud2} in 
Table~\ref{table:main_experiment_structured_reg_NYUd2}. 
\paragraph{Training details} We use ResNet-50 based encoder-decoder model for depth estimation \cite{hu2019revisiting} for all experiments. We train each model 20 epochs with batch size 32 and Adam optimizer. The learning rate is $1 \times 10^{-4}$. For last output layers of uncertainty estimation, the learning rate is one magnitude smaller for stable training. Following the benchmark \cite{yang2021delving}, the bin length is 0.1 meter. The many-shot region is defined as bins with over $2.6×10^7$ training pixels, the few-shot region are bins with fewer than $1.0 × 10^7$ training pixels, and other bins are within medium-shot region. We use $L_2$ loss for baselines and Laplacian negative log-likelihood loss for our proposed method.
\paragraph{Results} Compared to BalancedMSE and LDS,FDS, \MethodName performs better on the many-shot region and competitively on the medium shot region, while it is outperformed in the few-shot region. Although \MethodName achieves a smaller improvement over the Vanilla model in the few shot region, it preserves and actually improves the performance on the many-shot region. The competing approaches sacrifice that part of the distribution, and do perform significantly worse than the Vanilla model.

\begin{table}[t!]
\centering
\caption{\textbf{Experimental results for structured regression on the NYUd2 depth estimation dataset.} Regression performance RMSE$\downarrow$ .  
}
\label{table:main_experiment_structured_reg_NYUd2}
\begin{tabular}{lccccccccc}
\toprule
\multicolumn{1}{c}{} & \multicolumn{4}{c}{\textbf{NYUd2} $\downarrow$}  \\  \cmidrule{3-4} 
\multicolumn{1}{c}{}    & \multicolumn{1}{c}{\textbf{All}} & \multicolumn{1}{c}{\textbf{Many}} & \multicolumn{1}{c}{\textbf{Med.}} & \multicolumn{1}{c}{\textbf{Few}}  \\ \cmidrule{2-5}
Vanilla    &  1.48      & \underline{0.59} &  0.95  & 2.12 \\ \cmidrule{2-5}
INV &  \textbf{1.23} & 0.81 & 0.88 & \textbf{1.64}    \\ 
+LDS,FDS & 1.34  & 0.67 & \textbf{0.85} & 1.88 \\ \cmidrule{2-5}
BalancedMSE &  \underline{1.28} &0.79 &0.87 & \underline{1.74} \\ \cmidrule{2-5}  
\textbf{\MethodName} & 1.40 & \textbf{0.55}  & \underline{0.86} & 2.02  \\

\bottomrule

\end{tabular}
\end{table}

\end{document}